\documentclass[sn-mathphys,Numbered]{layout/sn-jnl}

\usepackage{graphicx}%
\usepackage{multirow}%
\usepackage{amsmath,amssymb,amsfonts}%
\usepackage{amsthm}%
\usepackage{mathrsfs}%
\usepackage[title]{appendix}%
\usepackage{xcolor}%
\usepackage{textcomp}%
\usepackage{manyfoot}%
\usepackage{booktabs}%
\usepackage{algorithm}%
\usepackage{algorithmicx}%
\usepackage{algpseudocode}%
\usepackage{listings}%
\usepackage[capitalize,noabbrev]{cleveref}

\usepackage{subfig}
\usepackage[algo2e]{algorithm2e}

\theoremstyle{thmstyleone}%
\theoremstyle{thmstyletwo}%

\theoremstyle{thmstylethree}%

\begin{document}

\title[Article Title]{Enhancing Monte Carlo Dropout Performance for Uncertainty Quantification}

\author*[1]{\fnm{Hamzeh} \sur{Asgharnezhad}} \email{h.asgharnezhad@deakin.edu.au}
\equalcont{These authors contributed equally to this work.}

\author[2]{\fnm{Afshar} \sur{Shamsi}}
\equalcont{These authors contributed equally to this work.}

\author[1]{\fnm{Roohallah} \sur{Alizadehsani}}

\author[2]{\fnm{Arash} \sur{Mohammadi}}
  
\author*[4]{\fnm{Hamid} \sur{Alinejad-Rokny}}

\affil[1]{\orgdiv{Institute for Intelligent Systems Research and Innovation (IISRI)}, \orgname{Deakin University}, \orgaddress{\street{Street}, \city{Geelong}, \postcode{3216}, \state{Victoria}, \country{Australia}}}

\affil*[2]{\orgdiv{Concordia Institute for Information Systems Engineering}, \orgname{Concordia University}, \orgaddress{ \city{Montréal}, \state{Québec}, \country{Canada}}}

\affil[4]{\orgdiv{The Graduate School of Biomedical Engineering}, \orgname{UNSW Sydney}, \orgaddress{\city{Sydney}, \country{Australia}}}

\abstract{
Knowing the uncertainty associated with the output of a deep neural network is of paramount importance in making trustworthy decisions, particularly in high-stakes fields like medical diagnosis and autonomous systems. Monte Carlo Dropout (MCD) is a widely used method for uncertainty quantification, as it can be easily integrated into various deep architectures. However, conventional MCD often struggles with providing well-calibrated uncertainty estimates. To address this, we introduce innovative frameworks that enhances MCD by integrating different search solutions namely Grey Wolf Optimizer (GWO), Bayesian Optimization (BO), and Particle Swarm Optimization (PSO) as well as an uncertainty-aware loss function, thereby improving the reliability of uncertainty quantification. We conduct comprehensive experiments using different backbones, namely DenseNet121, ResNet50, and VGG16, on various datasets, including Cats vs. Dogs, Myocarditis, Wisconsin, and a synthetic dataset (Circles). Our proposed algorithm outperforms the MCD baseline by 2–3\% on average in terms of both conventional accuracy and uncertainty accuracy while achieving significantly better calibration. These results highlight the potential of our approach to enhance the trustworthiness of deep learning models in safety-critical applications.

}


\maketitle

\section{Introduction}

Deep neural networks (DNNs) have become a cornerstone in fields such as medical diagnosis~\cite{esteva2017dermatologist, asgharnezhad2023improving}, drug discovery~\cite{chen2018rise}, and computer vision~\cite{krizhevsky2012imagenet, osowiechi2024watt}, owing to their ability to leverage vast datasets and advanced computational power. These networks, with their large number of parameters, excel in generalization tasks. However, a critical limitation of DNNs is their tendency to be overconfident in their predictions, even when they are incorrect. This overconfidence can lead to significant risks, particularly in high-stakes applications where decision-making accuracy is vital. To mitigate these risks, it is essential to accurately quantify the uncertainty associated with the predictions of DNNs~\cite{doan2024bayesian}. In the artificial intelligence literature, uncertainty is typically divided into two categories: epistemic and aleatoric~\cite{matthies2007quantifying}. Epistemic uncertainty, also known as model uncertainty, arises from the model’s limited knowledge and can be reduced by incorporating more data into the model. In contrast, aleatoric uncertainty, or data uncertainty, is inherent in the data due to factors such as noise and class overlap, and cannot be reduced even with additional data.

Traditionally, Bayesian neural networks have been the primary method for uncertainty quantification, providing a probabilistic framework by assigning distributions to model parameters and using Bayes' theorem to estimate uncertainties. However, the practical application of Bayesian methods is often hindered by their computational complexity, as they require solving high-dimensional integrals. Although approximation techniques like Markov Chain Monte Carlo (MCMC) techniques~\cite{neal2012bayesian}, Hamiltonian methods~\cite{springenberg2016bayesian}, and variational Bayesian techniques~\cite{blundell2015weight,graves2011practical} have been developed to address this issue, their effectiveness is limited by the quality of the approximations and the choice of priors~\cite{rasmussen2005healing}, making these methods challenging to implement in real-world scenarios. Recognizing the need for a more practical approach, Gal and Ghahramani~\cite{gal2016dropout} proposed Monte Carlo Dropout (MCD), which offers a Bayesian approximation by enabling dropout during both training and inference. MCD has gained popularity due to its simplicity and ease of integration into existing neural network architectures. However, despite its widespread use, MCD has been criticized for producing overly broad and unreliable uncertainty estimates, limiting its effectiveness in critical applications~\cite{coulston2016approximating}.

Building on these insights, this study introduces a novel framework that integrates an uncertainty-aware loss function with three advanced hyperparameter optimisation techniques, Grey Wolf Optimizer (GWO)~\cite{mirjalili2014grey}, Bayesian Optimisation (BO)~\cite{mockus2005bayesian}, and Particle Swarm Optimisation (PSO)~\cite{bonyadi2017particle}, to enhance the Monte Carlo Dropout (MCD) algorithm for improved predictive accuracy and uncertainty calibration. Unlike conventional MCD, which often produces poorly calibrated uncertainty estimates, our approach explicitly incorporates predictive entropy (PE) into the loss function, combining binary cross-entropy with a penalty term based on PE. This ensures that incorrect predictions exhibit high PE, reflecting high uncertainty, while correct predictions maintain low PE, indicating high confidence. This alignment between predictive confidence and classification accuracy improves the model’s ability to distinguish between reliable and uncertain predictions. Furthermore, we utilize different backbone architectures, including DenseNet121, ResNet50, and VGG16, to evaluate the robustness and generalisability of the proposed framework across diverse feature extraction settings. By combining the strengths of GWO, BO, and PSO to find the best set of hyperparameters, our framework ensures that both predictive accuracy and uncertainty calibration are systematically optimized, increasing uncertainty for incorrect predictions and reducing it for correct ones.

The rest of this paper is organized as follows:
The datasets used in this study have been explained in Section \ref{sec: dataset}. Section~\ref{Sec:RW} provides a brief review of the MCD algorithm and its application for uncertainty quantification in deep learning. In Section~\ref{Sec:PM}, we describe the proposed algorithm in detail. Experiments and results are provided in Section~\ref{Sec:SR}, followed by concluding remarks and avenues for future research in Section~\ref{Sec:Concl}.

\begin{figure}[t]
    \centering
    \includegraphics[scale=0.25]{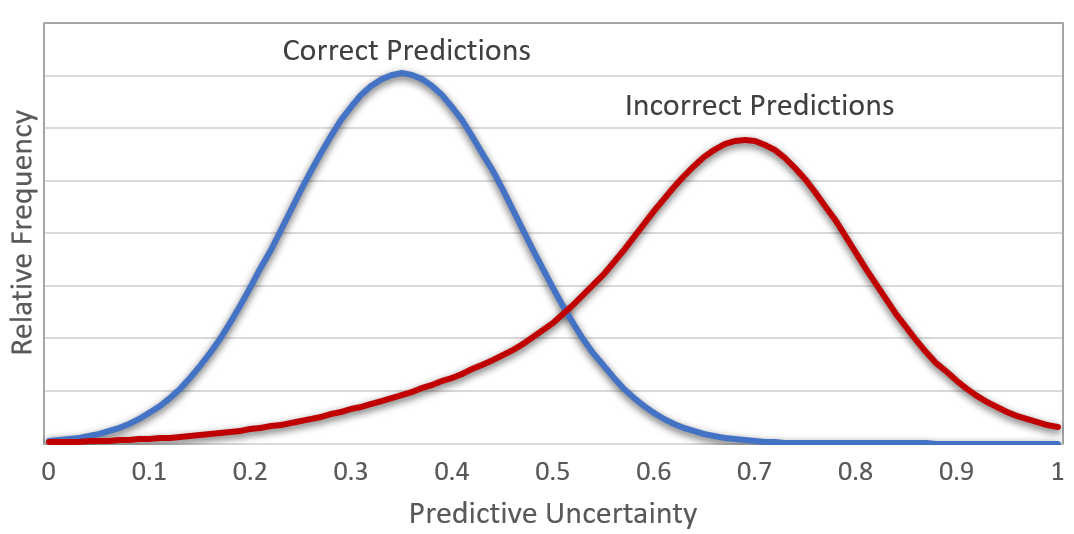}
    \caption{The predictions are classified to two groups. Red: predictions that are classified incorrectly and have high PE. Blue: predictions that are classified correctly and have low PE.}
    \label{Fig:Distributions}
\end{figure}

\section{Dataset}\label{sec: dataset}

To validate our result, two kind of datasets are used in this manuscript, namely Synthetic (Circles) and real (Myocardit, Cats vs Dogs, Wisconsin).
\subsection{Synthetic}

Circles is the sample toy dataset from Scikit Learn library that we will use 
to perform our results and benchmarking using different methods ($1000$ samples from each dataset)

\subsection{Myocardit~\cite{sharifrazi_myocarditis_dataset, sharifrazi2022cnn}}\label{sec: Myo}

The Myocardit dataset serves as the primary dataset and focal point of our study. Our analysis and conclusions will be largely derived from the data collected within this framework. Data collection for the Myocardit dataset was conducted at the CMR department of OMID Hospital in Tehran, Iran, from September 2018 to September 2019. The accuracy and integrity of the data collection process were supervised and approved by the local ethical committee of OMID Hospital.\\
Cardiac MRI (CMR) examinations were performed using a 1.5T system (MAGNETOM Aera, Siemens, Erlangen, Germany). All patients were scanned using dedicated body coils while positioned in the standard supine position. The CMR protocol included the following sequences: CINE-segmented images and pre-contrast T2-weighted (TRIM) images were obtained in both short- and long-axis views. Pre-contrast T1-weighted images were acquired in axial views of the myocardium, with the T1-weighted sequence immediately repeated following the injection of Gadolinium (DOTAREM 0.1 mmol/kg). Finally, Late Gadolinium Enhancement (LGE) images were acquired in high-resolution PSIR sequences, also in both short- and long-axis views.\\
The final dataset comprises 7,135 images/samples. It is important to note that working with real-world medical datasets presents challenges, particularly due to their imbalanced nature. In this context, one class may have a significantly higher density of samples than another, which can lead to the model being biased toward the more populous class during the training phase.

\subsection{Cats vs Dogs~\cite{microsoft_kaggle_cats_dogs}}

This dataset consists of a large collection of labeled images of cats and dogs, specifically designed for image classification tasks. It serves as a key resource for training and evaluating various image classification algorithms and models. Each image is annotated to indicate whether it depicts a cat or a dog, facilitating supervised learning approaches to accurately learn and distinguish between these two classes.

\subsection{Breast Cancer Wisconsin~\cite{street1993nuclear}}

The Breast Cancer Wisconsin Dataset is a well-known tabular dataset widely used for classification tasks, especially in the field of medical diagnostics. It consists of features derived from digitized images of fine needle aspirates of breast masses, with the primary objective of predicting whether a breast mass is malignant or benign based on these features. The dataset includes 569 instances, each characterized by 30 features, such as the mean, standard error, and worst (largest) values of various cell nucleus properties, including radius, texture, perimeter, area, smoothness, compactness, concavity, concave points, symmetry, and fractal dimension. Each instance is labeled as either malignant (indicating a cancerous mass) or benign (indicating a non-cancerous mass). Due to its relatively small size, balanced class distribution, and clinical relevance, the Wisconsin dataset is often used as a benchmark for evaluating the performance of machine learning algorithms, particularly in binary classification tasks.

\section{Related Works}\label{Sec:RW}
In the following subsections, we explain the fundamental concepts and background information required for this study.
\subsection{Monte Carlo Dropout}

In the artificial intelligence literature, dropout is primarily used as a regularization technique, encouraging neural networks to learn more robust patterns during training and preventing overfitting~\cite{srivastava2014dropout}. Gal and Ghahramani~\cite{gal2016dropout} demonstrated that the posterior distribution in a Bayesian setting can be approximated by performing multiple forward passes with dropout enabled during test time:
 
\begin{equation}\label{eq: mcd}
    \mu_{pred} \approx \frac{1}{T} \sum_t p(y = c \mid x, \hat\omega_t)
\end{equation}

\noindent where $x$ denotes the test input. $p(y = c \mid x, \hat\omega_t)$ represents the probability that $y$ belongs to $c$ (the output of softmax), and $\hat\omega_t$ denotes the model's parameters on the $t^{th}$ forward pass. $T$ shows the number of forward passes (MC iterations). Furthermore, they showed that Predictive Entropy (PE) can be used as a metric to estimate the associated uncertainty in a classification task which shows the degree of the relation of a prediction to each individual class.

\begin{equation}\label{Eq:MC-Dropout-PE}
    PE = - \sum_c \mu_{pred} \log \mu_{pred}
\end{equation}

\noindent where $c$ ranges over both classes. PE varies between $0$ and $1$. If PE is close to $1$, the prediction is highly uncertain.\\
The primary issue with Monte Carlo Dropout (MCD) is that the uncertainty estimates generated by this algorithm are not as well-calibrated compared to ensembles~\cite{lakshminarayanan2016simple}; largely because MCD's performance is heavily dependent on the dropout probability (dropout rate).Finding the optimal dropout rate is crucial, and this can be achieved using various search algorithms. However, optimization through search algorithms is typically constrained to models with a small number of parameters, making it essential to carefully select parameters based on their impact on the final prediction. To tackle this, Gal and Hron In~\cite{gal2017concrete}, the authors proposed a variant of dropout that can be tuned using gradient methods, enabling more calibrated uncertainty estimates for models with large numbers of parameters. 

\subsection{Uncertainty Accuracy (UA)}

Authors in~\cite{asgharnezhad2022objective} introduced the Uncertainty Confusion Matrix (UCM) and Uncertainty Accuracy (UA) metrics, which provide a means to objectively evaluate and compare different uncertainty quantification methods in a single dimension (noting that Bayesian methods typically work with distributions, which are generally two-dimensional). Uncertainty Accuracy (UAcc) is calculated as follows:

\begin{equation}\label{Eq:UAcc}
    UAcc \ = \ \frac{CC + IU}{\# samples}.
\end{equation}

\noindent CC shows correct and certain predictions and IU denotes incorrect and uncertain predictions. The larger the UA, the better the reliability of the uncertainties generated. UAcc is used to determine which algorithm performs better. They appear similar when visually inspected, but analysis reveals that they have different UAcc values. We select the algorithm with the higher UAcc as it will generate more reliable intervals.

\begin{table*}[t] 
\centering
\caption{The performance of the three different algorithms in the Circles in the presence of different noise levels in Circles dataset}
\label{tab:my-table1}
\resizebox{0.7\textwidth}{!}{%
    \begin{tabular}{llrrrr}
    \toprule
    Noise & Method & Accuracy & AUC & UAcc & ECE \\
    \midrule 
        & MCD & 96.50 & 96.50 & 91.00 & 7.20 \\
        & MCD plus PE & 96.50 & 96.50 & 91.50 & 5.44 \\
        0.05 & MCD plus GWO & \textbf{97.00} & \textbf{97.00} & 92.50 & \underline{3.37} \\
        & MCD plus BO & 96.50 & 96.50 & \textbf{93.00} & 3.69 \\
        & MCD plus PSO & 96.50 & 96.50 & \textbf{93.00} & \textbf{3.36} \\
    \midrule
        & MCD & 94.00 & 94.00 & 84.50 & 7.28 \\
        & MCD plus PE & \textbf{95.50} & \textbf{95.50} & 83.50 & 6.38 \\
        0.06 & MCD plus GWO & 94.50 & 94.50 & 87.50 & \underline{4.08} \\
        & MCD plus BO & 94.00 & 94.00 & 85.00 & 4.13 \\
        & MCD plus PSO & \underline{95.00} & \underline{95.00} & \textbf{88.00} & \textbf{3.24} \\     
    \midrule
        & MCD & 92.50 & 92.50 & 73.50 & 9.46 \\
        & MCD plus PE & \textbf{93.00} & \textbf{93.00} & 77.00 & 9.37 \\
        0.07 & MCD plus GWO & 92.50 & 92.50 & \textbf{85.50} & \textbf{4.54} \\
        & MCD plus BO & \textbf{93.00} & \textbf{93.00} & \underline{83.00} & \underline{5.02} \\
        & MCD plus PSO & 92.50 & 92.50 & 76.50 & 8.23 \\
    \bottomrule
    \end{tabular}
}
\end{table*}

\begin{table*}[t]
\centering
\caption{Qualitative comparison of different algorithms and their output distributions of the Circles with different noise levels. $\mu_1$ and $\mu_2$ are the centers of mis-classified and correctly-classified distributions and $Dist$ defines the distance between the two mentioned distributions.}
\label{tab:my-table2}
\resizebox{0.6\textwidth}{!}{%
    \begin{tabular}{llrrr}
    \toprule
    Dataset & Method & $\mu_1$ & $\mu_2$ & Distance \\
    \midrule
        & MCD & 0.262 & 0.585 & 0.323 \\
        & MCD plus PE & 0.246 & 0.588 & 0.342 \\
        0.05 & MCD plus GWO & 0.160 & 0.513 & 0.354 \\
        & MCD plus BO & 0.185 & 0.580 & \underline{0.395} \\
        & MCD plus PSO & 0.149 & 0.562 & \textbf{0.413} \\
    \midrule
        & MCD & 0.280 & 0.590 & 0.310 \\
        & MCD plus PE & 0.276 & 0.546 & 0.270 \\
        0.06 & MCD plus GWO & 0.141 & 0.517 & \textbf{0.376} \\
        & MCD plus BO & 0.225 & 0.563 & 0.338 \\
        & MCD plus PSO & 0.154 & 0.496 & \underline{0.342} \\
    \midrule
        & MCD & 0.398 & 0.592 & 0.194 \\
        & MCD plus PE & 0.386 & 0.584 & 0.198 \\
        0.07 & MCD plus GWO & 0.228 & 0.492 & \underline{0.264} \\
        & MCD plus BO & 0.257 & 0.525 & \textbf{0.268} \\
        & MCD plus PSO & 0.362 & 0.609 & 0.247 \\
    \bottomrule
    \end{tabular}
}
\end{table*}

\subsection{Expected Calibration Error}

It is essential to assess how well the predictions of a deep neural network are calibrated. The concept of Expected Calibration Error (ECE) was introduced in~\cite{guo2017calibration}. To calculate ECE, predictions are first grouped into different bins (denoted as M bins) based on their softmax output. The ECE for each bin is then determined by calculating the difference between the fraction of correctly classified predictions and the mean confidence (probability) for that bin. The overall ECE is obtained by taking the weighted average of these errors across all bins:

\begin{equation}\label{Eq:ECE}
    ECE = \sum_{m=1}^{M} \frac{|B_m|}{n} \left | acc(B_m) - conf(B_m) \right |
\end{equation}

\noindent where $ acc(B_m) $ and $ conf(B_m)$ are the accuracy and confidence for the $m^{th}$ bin:
\begin{equation}\label{Eq:ECE-Acc}
    acc(B_m) = \sum \frac{1}{|B_m|} \textbf{1} \left (\hat{y_i} = y_i \right )
\end{equation}

\begin{equation}\label{Eq:ECE-Conf}
    conf(B_m) = \sum \frac{1}{|B_m|} p_i
\end{equation}

\noindent where $\textbf{1}(\cdot)$ is the indicator function.

\section{Proposed Method}\label{Sec:PM}

\begin{figure}[t]
    \centering    
    \begin{minipage}[b]{0.45\textwidth}
        \centering
        \subfloat[Week preformed model]{\includegraphics[width=\textwidth]{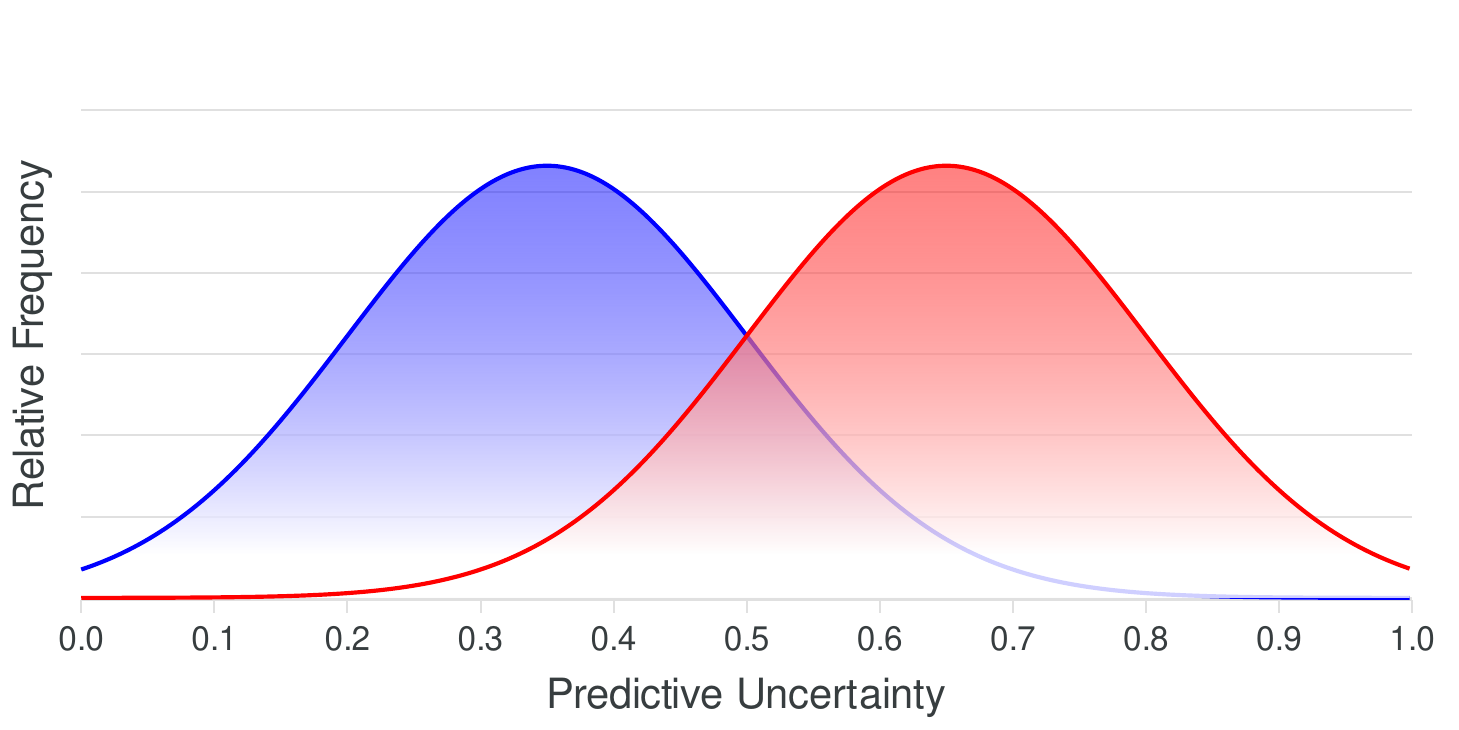}}    
    \end{minipage}
    \hfill
    \begin{minipage}[b]{0.45\textwidth}      
        \centering
        \subfloat[Well preformed model]{\includegraphics[width=\textwidth]{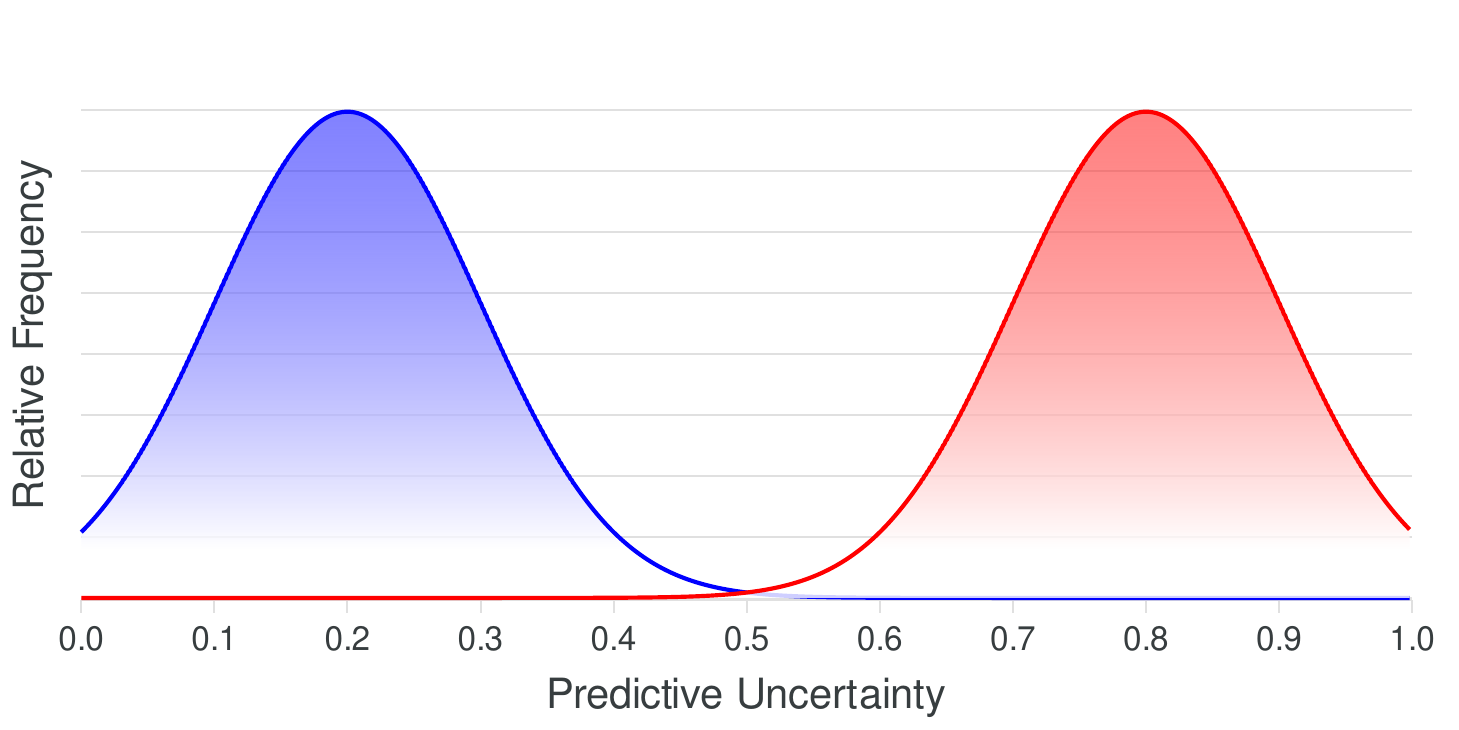}}
    \end{minipage}
    \caption{Comparison of two models’ perspectives on uncertainty. The predictions are classified into two groups. Red: predictions that are classified incorrectly and have high PE. Blue: predictions that are classified correctly and have low PE.}
    \label{Fig:Distributions_Comparison}
\end{figure}

In a classification task, the final predictions of a deep neural network can be categorized into two main groups based on their Predictive Entropy (PE): correctly classified and misclassified, as illustrated in Fig.~\ref{Fig:Distributions}. Ideally, incorrect predictions should exhibit high PE, indicating high uncertainty, while correct predictions should have low PE, reflecting low uncertainty and high confidence. As previously discussed, high entropy signifies high uncertainty, while low entropy indicates low uncertainty and high confidence for the predicted samples in the test dataset. The distributions of correctly classified (blue) and misclassified (red) predictions are shown based on sorted predictive entropies. A well-designed model should exhibit uncertainty when it makes mistakes, meaning it should recognize when it is unsure about a decision. Conversely, a robust model should display confidence when it makes accurate predictions, clearly indicating the certainty of its correct judgments.\\
In the PE distribution of an ideal model, we would expect high uncertainty (close to one) for misclassified data (indicated by the red color) and low uncertainty (close to zero) for correctly classified data (indicated by the blue color).  Fig.~\ref{Fig:Distributions_Comparison} shows two examples of PE distribution for a well-performing model and a poorly performing model. 

Building on this understanding of predictive entropy (PE) and its relationship to uncertainty in model predictions, it is critical to address the calibration of uncertainty measures, as highlighted in \cite{lakshminarayanan2017simple}. In particular, the uncertainty captured by MCD algorithms is often poorly calibrated, which limits the model's ability to effectively distinguish between confident and uncertain predictions. To overcome this limitation, we propose an enhanced approach that integrates uncertainty accuracy into the model's optimization process. Specifically, we introduce the following loss function: 

\begin{align}
\hat{\mu}^{pred}_{b,c} &= \frac{1}{M} \sum_{m=1}^{M} \hat{y}^{(m)}_{b,c}, b=1:B, c=1:C, \label{mu_pred_bc_eq}\\
\hat{\mu}^{pred}_{b} &= \underset{c}{\mathrm{argmax}}\:\: \hat{\mu}^{pred}_{b,c}, \label{mu_pred_b_eq}\\
Loss &= \frac{1}{B} \sum_{b=1}^{B} \left \lbrace
\underbrace{ -\left[y_b \log \left( \hat{\mu}^{pred}_{b} \right) + (1 - y_b) \log \left( 1- hat{\mu}^{pred}_{b} \right) \right]}_{Binary \: cross \: entropy} + \sum_{m=1}^{M} PE(x_{b}^{(m)}) \right\rbrace
,\label{Eq:LF-PE}
\end{align}
where $M$ represents the number of Monte Carlo Dropout (MCD) forward passes, $B$ denotes the batch size, and $C$ corresponds to the number of classes. The variable $y_b$ signifies the target label for the input sample $x_b$, while $\hat{y}^{(m)}_{b,c}$ is a $[B \times C]$ matrix, where the $b$th row contains the network’s softmax output prediction for $x_b$. The left-hand side of Equation \ref{mu_pred_bc_eq} is a $[B \times C]$ matrix, and the left-hand side of Equation \ref{mu_pred_b_eq} is a $[B \times 1]$ vector due to the application of the argmax operator. Finally, $PE(x_b^{(m)})$ represents the predictive entropy for $x_b$ in the $m$th MCD forward pass.

This formulation not only aligns the optimization process with the primary objective of improving predictive accuracy but also directly incorporates uncertainty calibration into the training process by penalizing predictions with poorly calibrated entropy values. 
The above loss function can be served as a fitness function for any hyperparameter optimization algorithm (e.g., Bayesian optimization, evolutionary strategies, or gradient-based methods), guiding the search process toward configurations that balance predictive accuracy and uncertainty calibration.
To enable the optimization of hyperparameters alongside model weights, we treat hyperparameters as additional variables within a broader search space. This is because some hyperparameters such as dropout rate, which has a great impact on the output calibration of the MCD, will not be optimized during training as such hyperparametr optimization will help the model to find the best configurations for these.

 In our framework, hyperparameters such as, dropout rate, or the number of neurons in the hidden layers are explicitly included in the optimization routine. By iteratively minimizing the proposed loss, the model not only adjusts its weights but also identifies hyperparameter configurations that yield better-calibrated uncertainty estimates. This dual-level optimization ensures that both the model's predictions and its associated uncertainties are systematically improved, ultimately resulting in a more confident and reliable uncertainty-aware model. In the following subsection, we investigate the impact of this dual-level optimization approach using several well-known hyperparameter optimization algorithms, namely Grey Wolf Optimizer (GWO), Bayesian Optimization (BO), and Particle Swarm Optimization (PSO). By employing these algorithms, we aim to evaluate their effectiveness in identifying hyperparameter configurations that enhance both predictive accuracy and uncertainty calibration within the proposed framework.

\subsection{Grey Wolf Optimizer (GWO)~\cite{mirjalili2014grey} }

As stated earlier, the calibration of predictions in MCD is highly dependent on the magnitude of hyperparameters, such as the dropout rate, which is defined by the user at the start of training. However, in our case, we aim to determine the optimal hyperparameter values using an additional search algorithm. The steps for utilizing GWO in this process are outlined in the following paragraphs.

To solve this optimization problem defined in Eq. (7-9), we employ the GWO, which models the social hierarchy and hunting behavior of grey wolves. In GWO, the positions of wolves in the search space correspond to the hyperparameter configurations. The best solution (leader) is represented as \( \boldsymbol{X}_{\alpha} \), while the second and third best solutions are denoted as \( \boldsymbol{X}_{\beta} \) and \( \boldsymbol{X}_{\delta} \), respectively. The rest of the wolves (search agents) update their positions iteratively according to:

\begin{align}
A &= 2a \times r_1 - a \\
C &= 2 \times r_2 \\
\boldsymbol{X}^{j}_{\text{new}} &= \boldsymbol{X}_{\alpha} - A \times |C \times \boldsymbol{X}_{\alpha} - \boldsymbol{X}^{j}_{\text{old}}|
\end{align}

where \( \boldsymbol{X}^{j}_{\text{old}} \) represents the position of the \( j^{th} \) search agent (hyperparameter set) in the current iteration, while \( \boldsymbol{X}_{\alpha} \) is the best solution found so far. The parameter \( a \) decreases linearly from 2 to 0 over the course of optimization, and \( r_1, r_2 \) are random values sampled from \([0,1]\). In our case, the values of \( \boldsymbol{X}^{j}_{\text{new}} \) correspond to the hyperparameters \( \lambda_j \). Thus, each hyperparameter's position is updated iteratively based on its current value and the best-found configuration using Eq. (11-13).

By iteratively optimizing \( \lambda_j \) over multiple iterations, GWO enables the discovery of hyperparameter configurations that improve uncertainty calibration while maintaining high predictive accuracy. This dual-level optimization ensures that both model parameters and hyperparameters contribute to achieving well-calibrated uncertainty estimates, making the model more robust for safety-critical applications such as autonomous driving and medical diagnosis.

\begin{figure*}[!t]
    \centering
    \begin{minipage}[b]{.28\textwidth}
        \subfloat[Circles with Noise = 0.05]{\includegraphics[width=\textwidth]{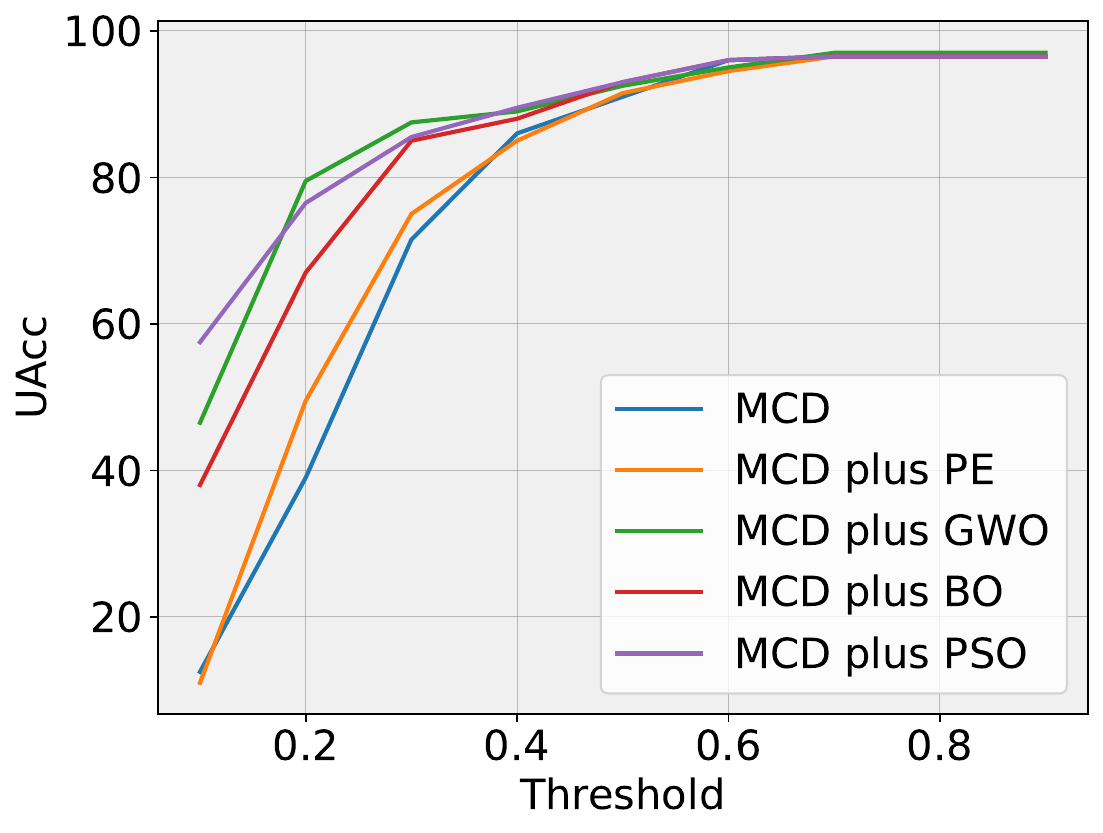}}
    \end{minipage}\qquad
    \begin{minipage}[b]{.28\textwidth}
        \subfloat[Circles with Noise = 0.06]{\includegraphics[width=\textwidth]{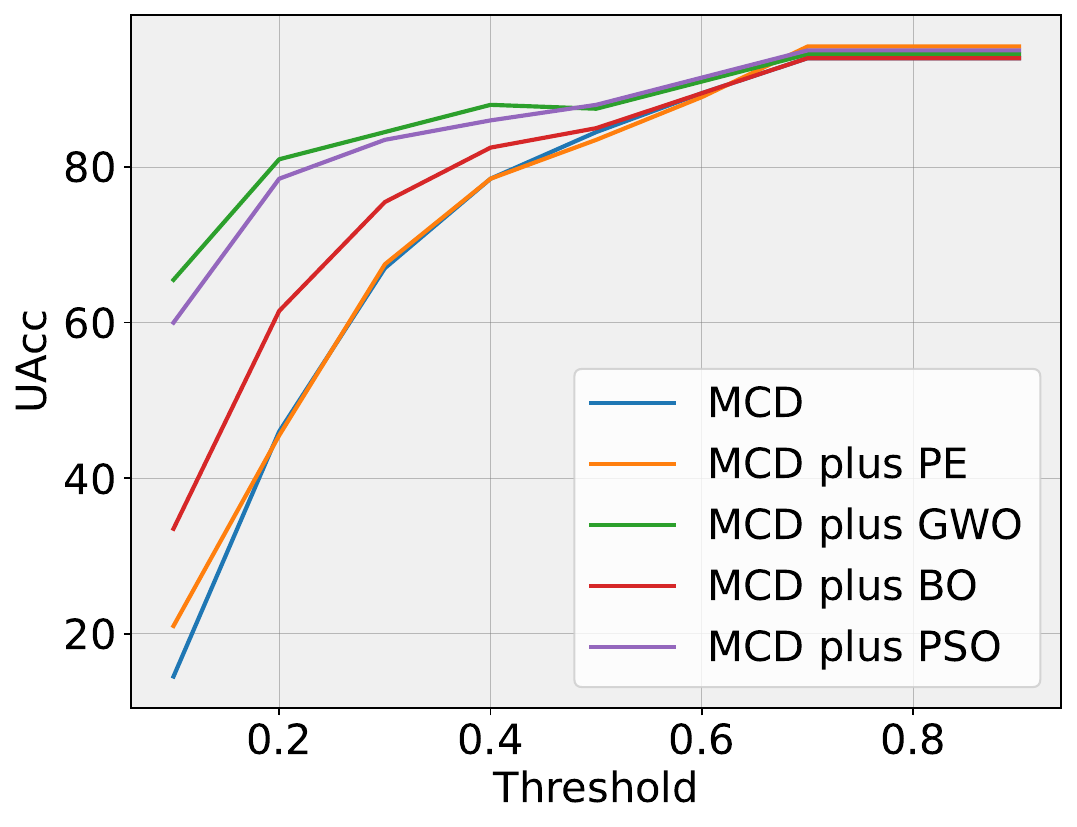}}
    \end{minipage}\qquad
    \begin{minipage}[b]{.28\textwidth}
        \subfloat[Circles with Noise = 0.07]{\includegraphics[width=\textwidth]{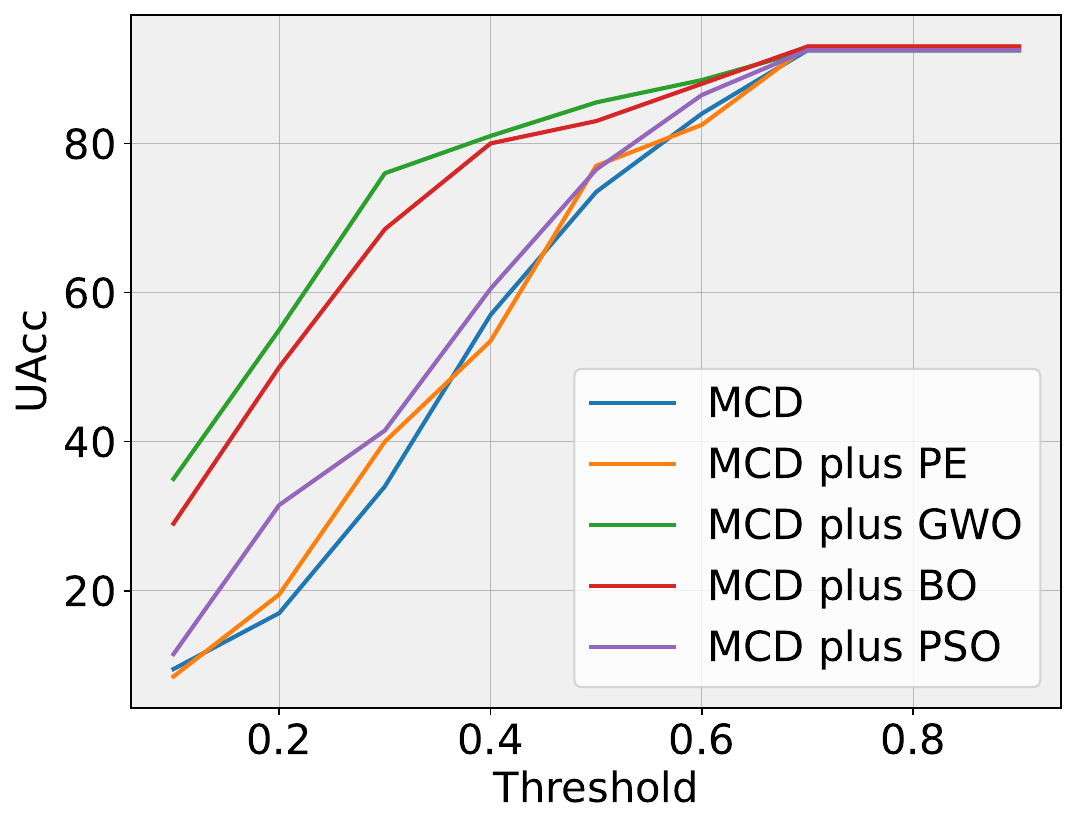}}
    \end{minipage}
    \caption{UAcc metric for three different algorithms with different noise levels (three different uncertainty levels). UAcc is calculated for different uncertainty thresholds.}
    \label{Fig:UAs}
\end{figure*}

\begin{figure}[t]
    \centering   \includegraphics[width=\textwidth]{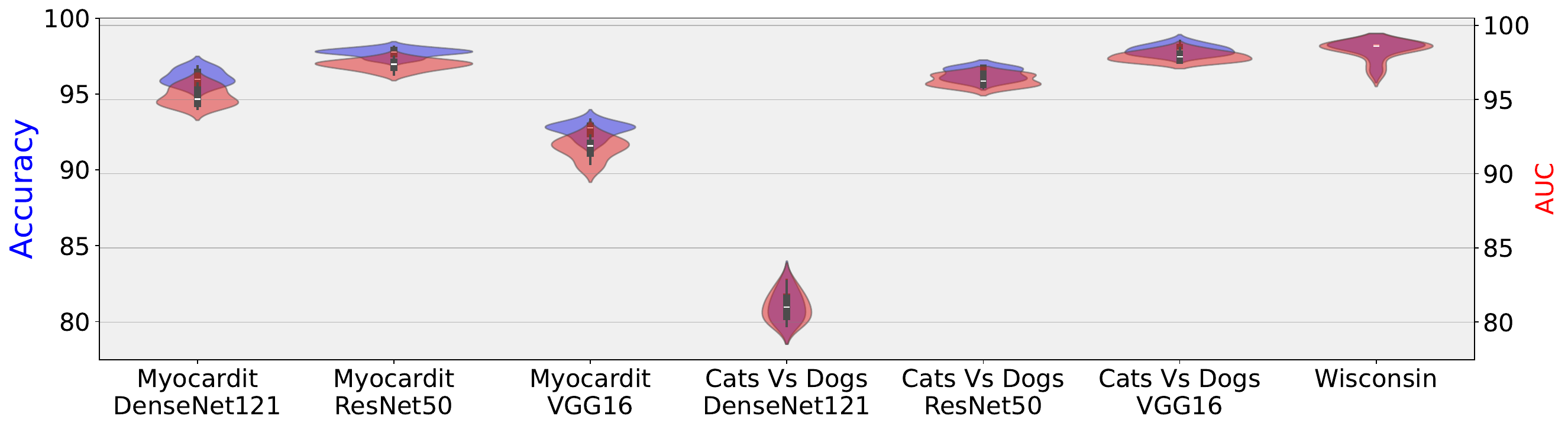}
    \caption{AUC and Accuracy plotted for real datasets,  Result generated by 10 times random initialization of wight, Training and testing.}
    \label{Fig: AccAuc}
\end{figure}

\subsection{Bayesian Optimizer (BO)~\cite{mockus2005bayesian}}

To find the best set of hyperparameters for MCD, BO could also be an alternative solution to systematically search for the optimal values that balance predictive accuracy and uncertainty calibration. Unlike heuristic-based approaches, BO employs a probabilistic model to approximate the objective function, allowing it to efficiently explore the search space and converge toward an optimal configuration. BO formulates hyperparameter tuning as an optimisation problem, where the goal is to minimise the loss function defined in Eq. (7-9):

\begin{equation}
\lambda^* = \underset{\lambda}{\text{arg min}} \quad \mathcal{L}(\lambda)
\end{equation}

\noindent where \( \lambda \) represents the hyperparameter set (e.g., dropout rate), and \( \mathcal{L}(\lambda) \) is the objective function measuring prediction calibration and accuracy. Since directly evaluating \( \mathcal{L}(\lambda) \) is computationally expensive, BO models it using a Gaussian Process (GP):

\begin{equation}
\mathcal{L}(\lambda) \sim GP(\mu(\lambda), k(\lambda, \lambda'))
\end{equation}

\noindent where \( \mu(\lambda) \) is the mean function representing the expected loss, and \( k(\lambda, \lambda') \) is the covariance function capturing relationships between different hyperparameter configurations. BO selects the next hyperparameter candidate by maximising an acquisition function:

\begin{equation}
\lambda_{t+1} = \underset{\lambda}{\text{arg max}} \quad \alpha(\lambda | D_t)
\end{equation}

\noindent where \( D_t = \{(\lambda_i, \mathcal{L}(\lambda_i))\}_{i=1}^{t} \) represents previous observations, and \( \alpha(\lambda) \) (e.g., Expected Improvement, Upper Confidence Bound) determines the next evaluation point. First, the objective function \( \mathcal{L}(\lambda) \) is evaluated for an initial set of hyperparameter configurations, typically using a space-filling design such as Latin Hypercube Sampling. Second, a GP model is fitted to the collected evaluations to approximate \( \mathcal{L}(\lambda) \). Third, the next hyperparameter configuration \( \lambda_{t+1} \) is selected by optimising the acquisition function \( \alpha(\lambda) \), which balances exploration and exploitation. Fourth, the model is trained with \( \lambda_{t+1} \), and the loss \( \mathcal{L}(\lambda_{t+1}) \) is computed and added to the dataset \( D_{t+1} \). Finally, these steps are repeated iteratively until convergence or a predefined stopping criterion is met.

Unlike traditional search methods, BO efficiently navigates the hyperparameter space by leveraging probabilistic models, requiring significantly fewer evaluations to find an optimal configuration. 

\subsection{Particle Swarm Optimisation (PSO)~\cite{bonyadi2017particle}}

To optimise the hyperparameters of MCD, Particle Swarm Optimisation (PSO) is another efficient approach that can be utilised. PSO is a population-based metaheuristic inspired by the collective behaviour of bird flocks or fish schools, where individuals (particles) explore the search space by iteratively adjusting their positions based on personal experiences and social interactions. Unlike Bayesian Optimisation, which builds a probabilistic model of the objective function, PSO relies on velocity-based updates to navigate the hyperparameter space effectively. The hyperparameter tuning problem in Eq. (7-9) can be formulated as an optimisation problem:

\begin{equation}
\lambda^* = \underset{\lambda}{\text{arg min}} \quad \mathcal{L}(\lambda)
\end{equation}

\noindent where \( \lambda \) represents the hyperparameter set (e.g., dropout rate), and \( \mathcal{L}(\lambda) \) is the objective function that measures the calibration and accuracy of predictions. In PSO, a set of particles (hyperparameter configurations) explore the search space, updating their positions based on their personal best solution \( \boldsymbol{p}_i \) and the global best solution \( \boldsymbol{g} \) found so far.

Each particle’s position and velocity are updated according to:

\begin{align}
v_i^{(t+1)} &= \omega v_i^{(t)} + c_1 r_1 (\boldsymbol{p}_i - \lambda_i^{(t)}) + c_2 r_2 (\boldsymbol{g} - \lambda_i^{(t)}) \\
\lambda_i^{(t+1)} &= \lambda_i^{(t)} + v_i^{(t+1)}
\end{align}

\noindent where \( v_i^{(t)} \) is the velocity of the \( i^{th} \) particle at iteration \( t \), \( \omega \) is the inertia weight controlling the balance between exploration and exploitation, \( c_1 \) and \( c_2 \) are acceleration coefficients determining the influence of personal and global best solutions, and \( r_1, r_2 \) are random numbers sampled from \([0,1]\).

First, an initial population of particles is generated, each representing a different hyperparameter configuration. Second, the objective function \( \mathcal{L}(\lambda) \) is evaluated for all particles to determine their fitness. Third, each particle updates its personal best solution \( \boldsymbol{p}_i \), and the global best \( \boldsymbol{g} \) is determined from the best-performing particle. Fourth, the velocity and position of each particle are updated using Eq. (15-16), ensuring movement toward promising solutions while maintaining diversity in exploration. Finally, these steps are iterated until convergence or a predefined stopping criterion is met.

Compared to other optimisation techniques, PSO offers a balance between exploration and exploitation, making it well-suited for non-convex optimisation problems like hyperparameter tuning.

\section{Simulations and Results}\label{Sec:SR}

This section is divided into two main parts. The first subsection presents the results obtained from the synthetic dataset, followed by the results for the real-world datasets.

\begin{figure*}[!t]
\centering
    \begin{minipage}[b]{.28\textwidth}
        \subfloat[Myocardit, VGG16]{\includegraphics[width=\textwidth]{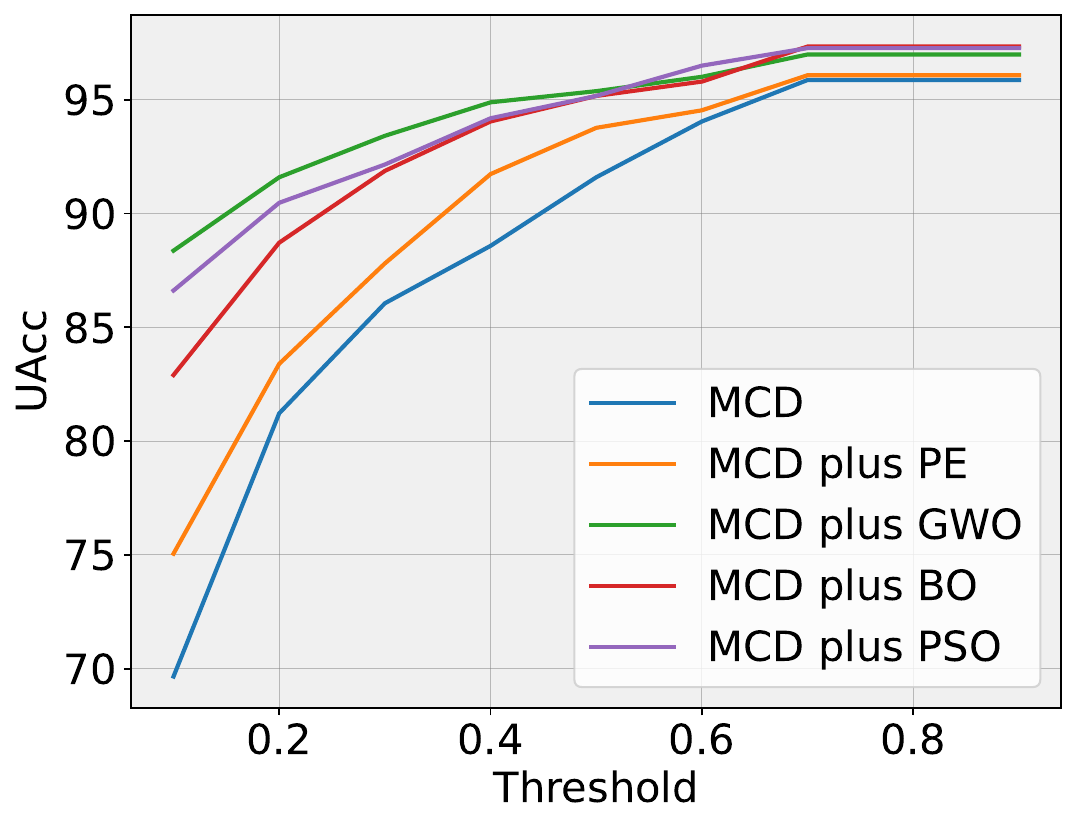}}
    \end{minipage}\qquad
    \begin{minipage}[b]{.28\textwidth}
        \subfloat[Myocardit, ResNet50]{\includegraphics[width=\textwidth]{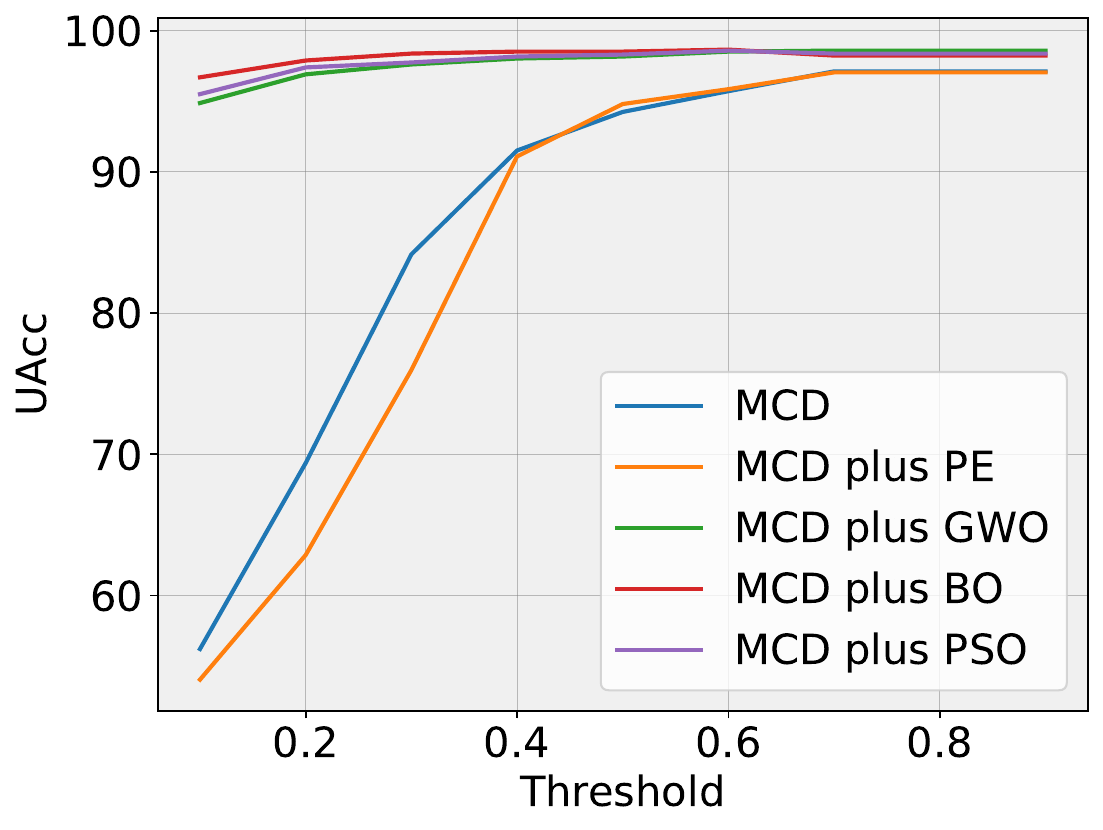}}
    \end{minipage}
    \begin{minipage}[b]{.28\textwidth}
        \subfloat[Myocardit, DenseNet121]{\includegraphics[width=\textwidth]{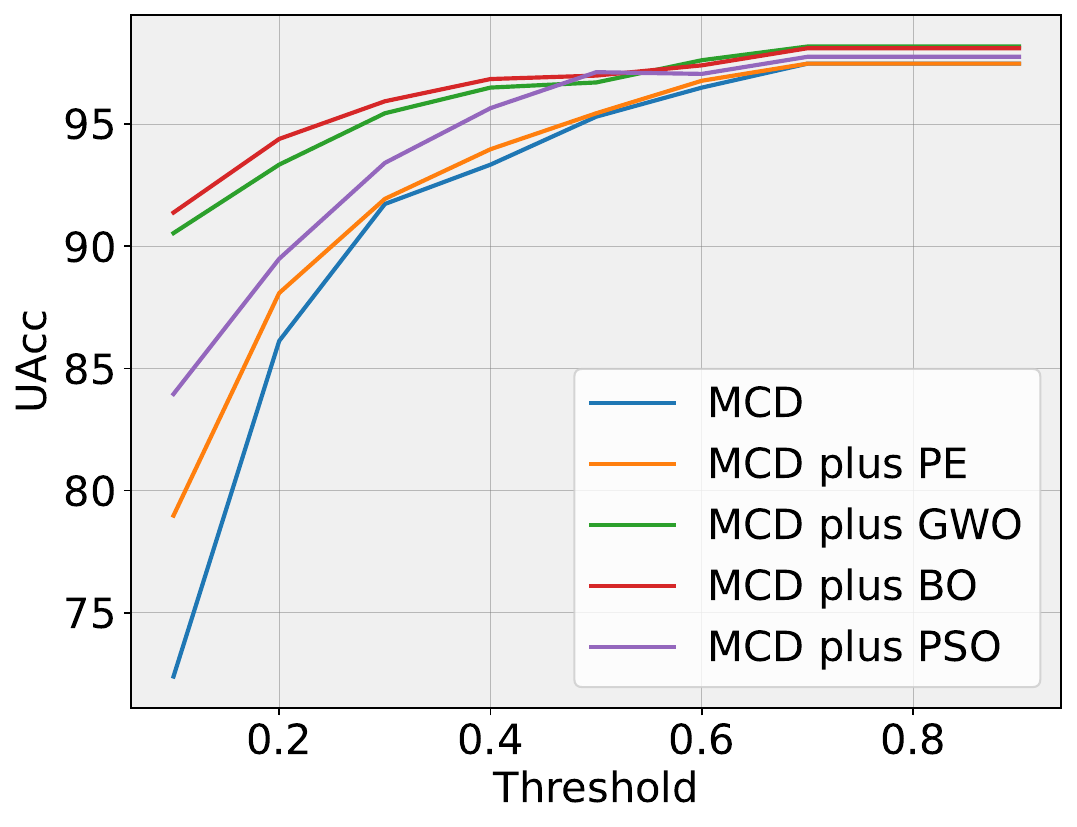}}
    \end{minipage}\qquad 
    \begin{minipage}[b]{.28\textwidth}
        \subfloat[Cats Vs Dogs, VGG16]{\includegraphics[width=\textwidth]{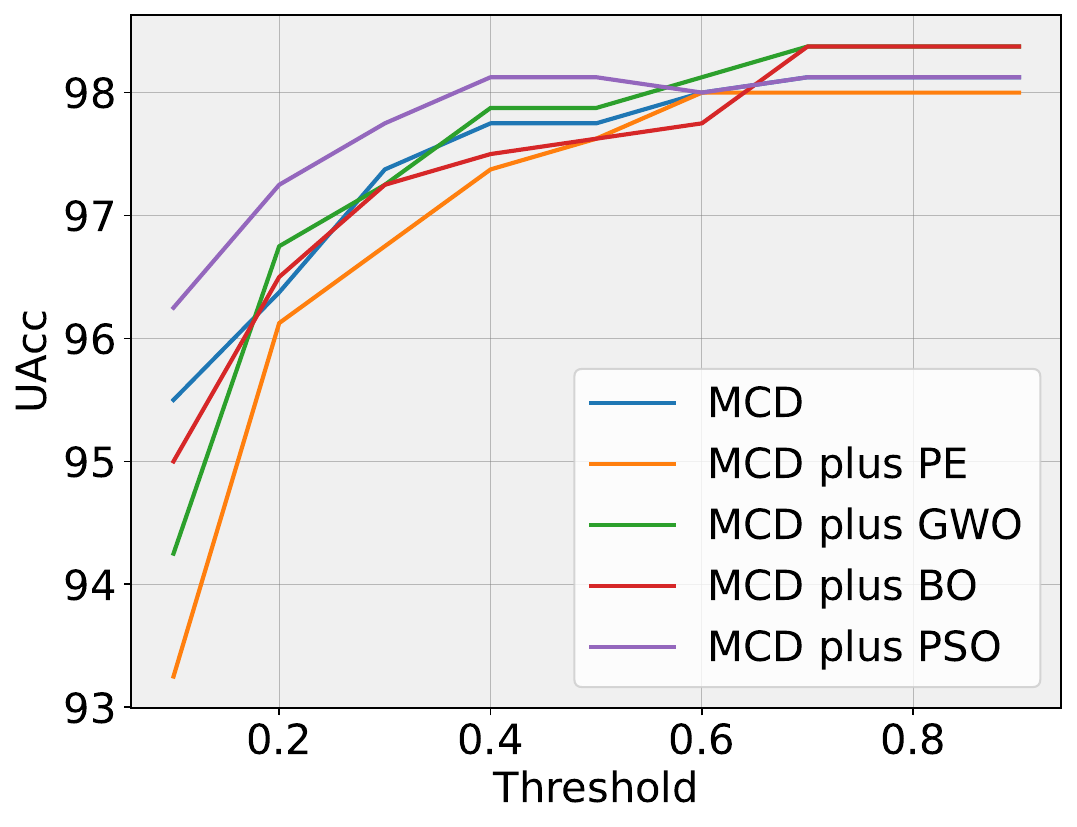}}
    \end{minipage}\qquad
    \begin{minipage}[b]{.28\textwidth}
        \subfloat[Cats Vs Dogs, ResNet50]{\includegraphics[width=\textwidth]{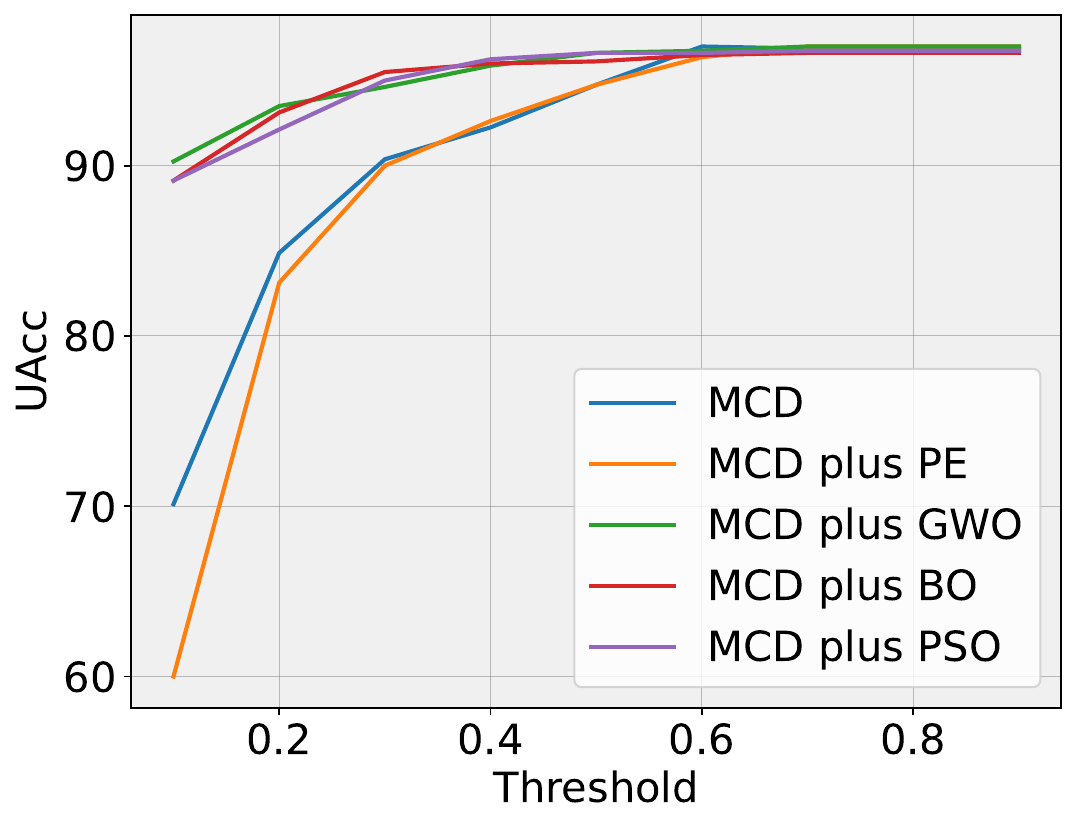}}
    \end{minipage}
    \begin{minipage}[b]{.28\textwidth}
        \subfloat[Cats Vs Dogs, DenseNet121]{\includegraphics[width=\textwidth]{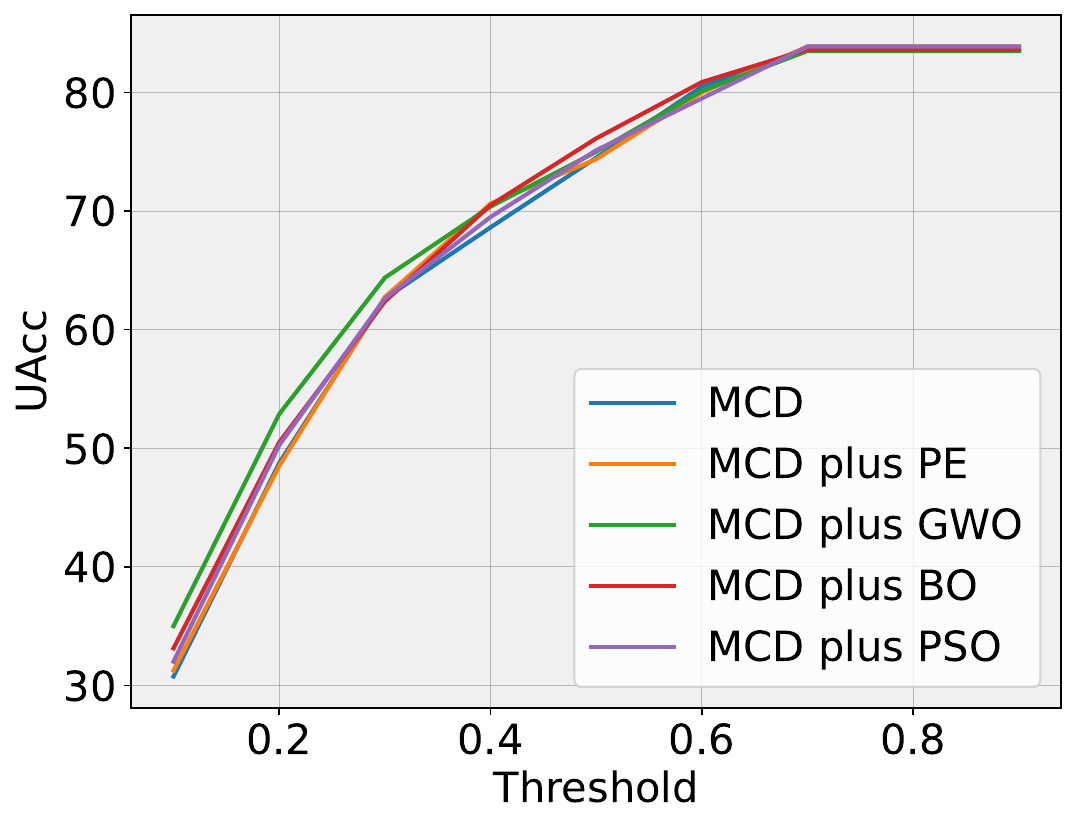}}
    \end{minipage}\qquad 
    \begin{minipage}[b]{.28\textwidth}
        \subfloat[Wisconsin]{\includegraphics[width=\textwidth]{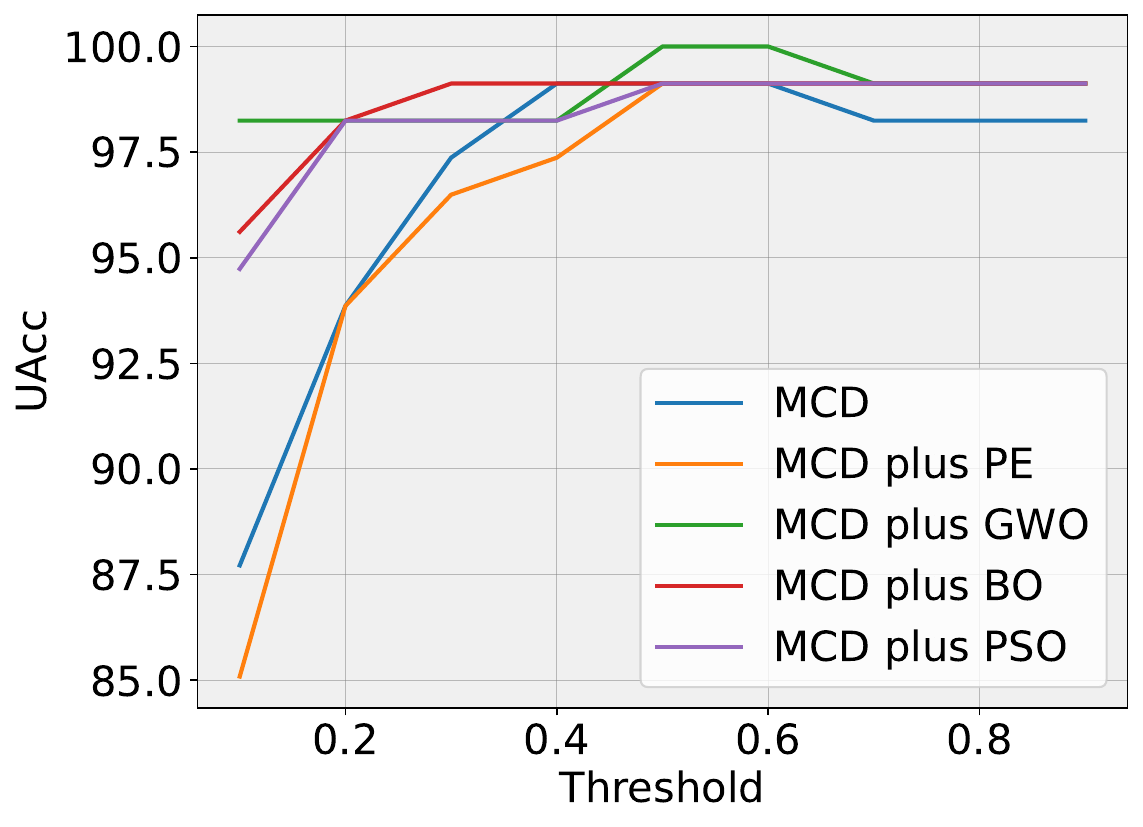}}
    \end{minipage}    
    \caption{The UAcc of five different algorithms for different thresholds are shown for different datasets. All suggested solutions outperform base MCD in terms of capturing better uncertainty.}
    \label{Fig: Myocardit}
\end{figure*}

\begin{figure}[t]
    \centering   \includegraphics[width=\textwidth]{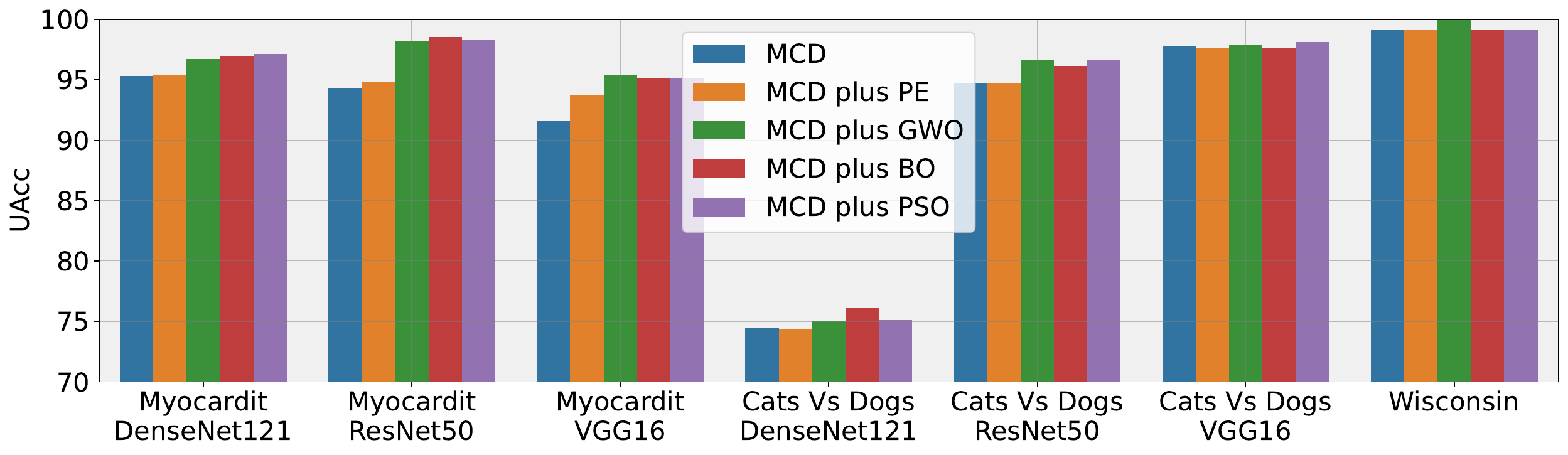}
    \caption{Uncertainty Accuracy for Threshold = 0.5}
    \label{Fig: UAThr}
\end{figure}

\begin{figure*}[!t]
    \centering
    \begin{minipage}[b]{.45\textwidth}
        \subfloat[Myocardit, VGG16]{\includegraphics[width=\textwidth]{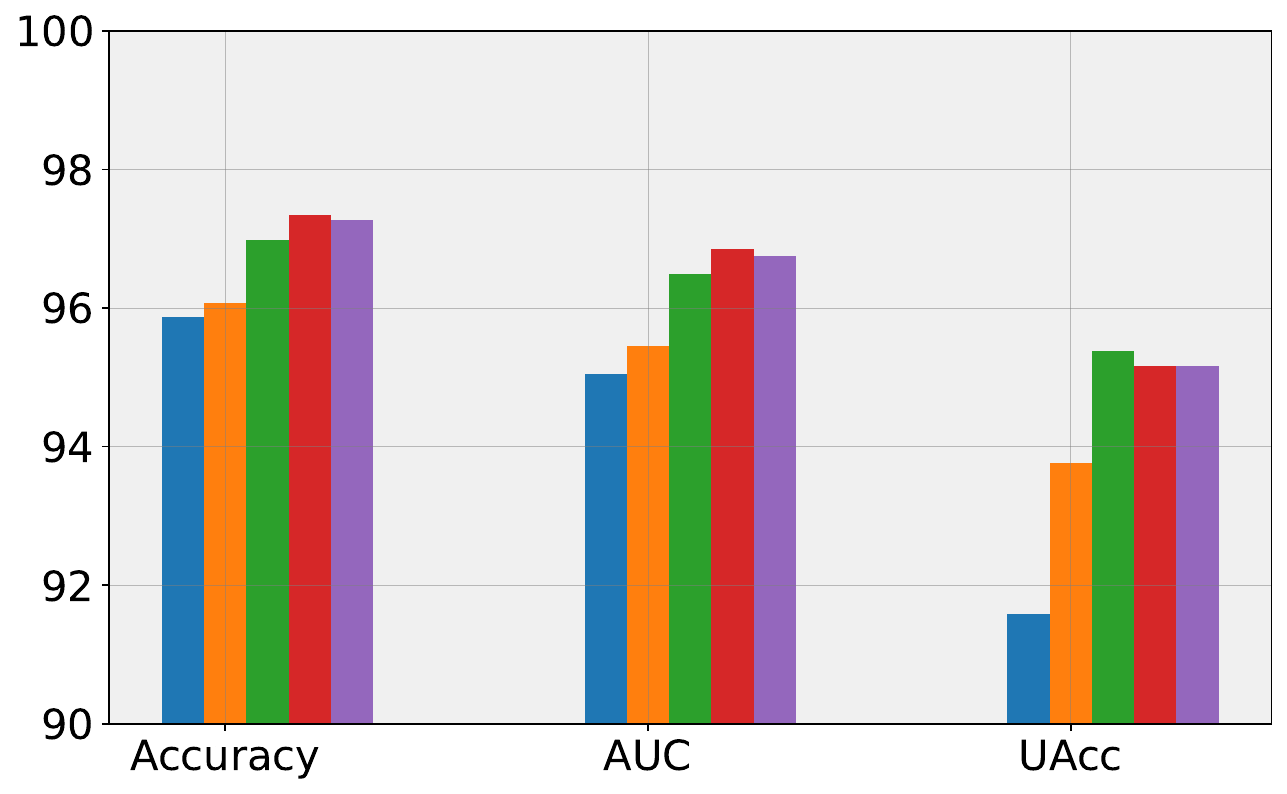}}
    \end{minipage}  
    \begin{minipage}[b]{.45\textwidth}
        \subfloat[Myocardit, ResNet50]{\includegraphics[width=\textwidth]{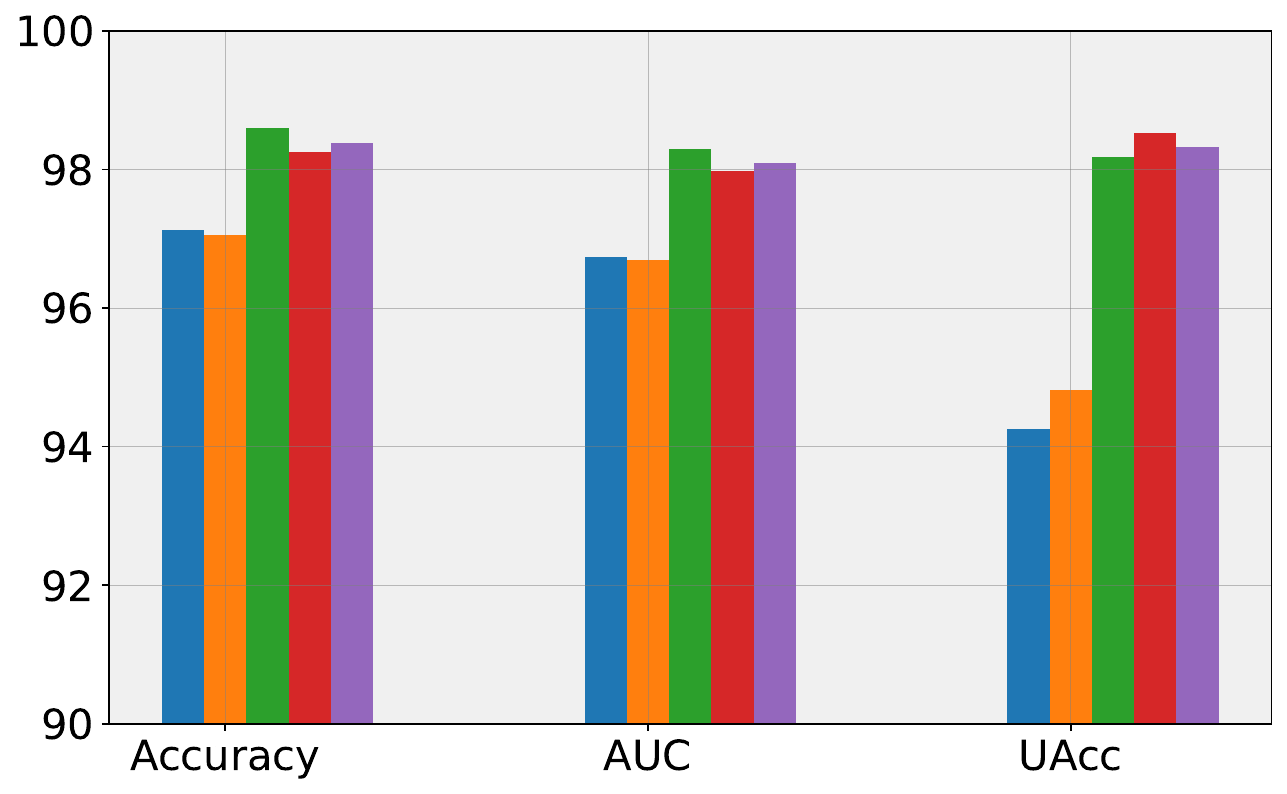}}
    \end{minipage}  
    \begin{minipage}[b]{.45\textwidth}
        \subfloat[Myocardit, DenseNet121]{\includegraphics[width=\textwidth]{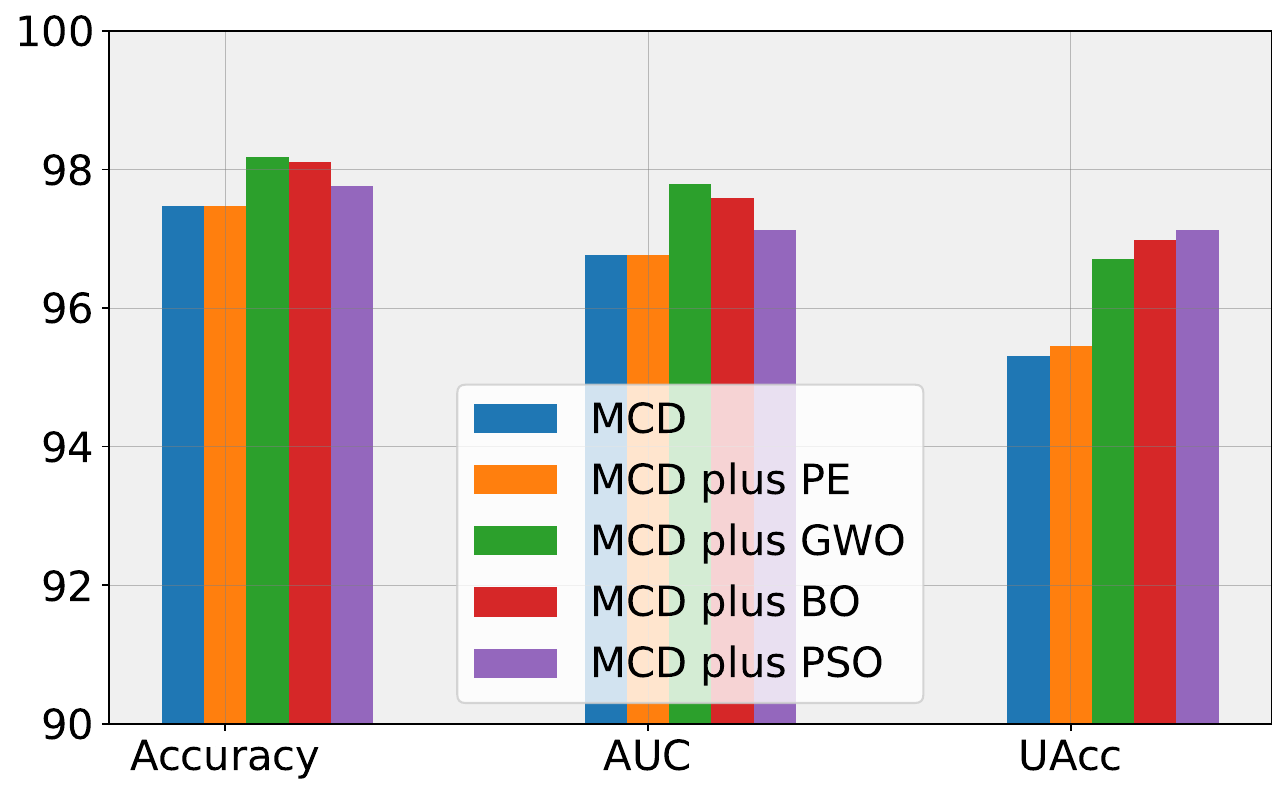}}
    \end{minipage}  
    \caption{Comparison of three metrics (Accuracy, AUC, UAcc across different methods used in the study for Myocardit as our main dataset.}
    \label{Fig: com3met}
\end{figure*}

\begin{table*}[t]
    \centering
    \caption{Optimised parameters for different algorithm on different dataset. it should be noted that Wisconsin is a tabular datset and feature extractor dose not apply to it.} 
    \label{tab:optimisedparams}
    \resizebox{0.8\textwidth}{!}{%
    \begin{tabular}{lllrrll}
    \toprule
    Dataset & Feature extractor & Optimizer & L1 & L2 & P1 & P2 \\  
    \midrule
     & DenseNet121 & BO & 229 & 50 & 0.400 & 0.170 \\
     & DenseNet121 & GWO & 237 & 38 & 0.220 & 0.390 \\
     & DenseNet121 & PSO & 215 & 31 & 0.500 & 0.240 \\
     & ResNet50 & BO & 220 & 39 & 0.270 & 0.480 \\
    Myocardit & ResNet50 & GWO & 255 & 46 & 0.240 & 0.490 \\
     & ResNet50 & PSO & 217 & 29 & 0.150 & 0.120 \\
     & VGG16 & BO & 237 & 38 & 0.220 & 0.390 \\
     & VGG16 & GWO & 241 & 45 & 0.140 & 0.350 \\
     & VGG16 & PSO & 224 & 49 & 0.170 & 0.510 \\
    \midrule
     & DenseNet121 & BO & 220 & 42 & 0.370 & 0.240 \\
     & DenseNet121 & GWO & 255 & 45 & 0.270 & 0.370 \\
     & DenseNet121 & PSO & 214 & 39 & 0.360 & 0.250 \\
     & ResNet50 & BO & 223 & 34 & 0.240 & 0.420 \\
     Cats Vs Dogs & ResNet50 & GWO & 256 & 50 & 0.150 & 0.440 \\
     & ResNet50 & PSO & 210 & 36 & 0.130 & 0.240 \\
     & VGG16 & BO & 236 & 44 & 0.370 & 0.330 \\
     & VGG16 & GWO & 246 & 53 & 0.170 & 0.330 \\
     & VGG16 & PSO & 223 & 29 & 0.160 & 0.440 \\  
    \midrule
     & - & BO & 216 & 38 & 0.220 & 0.390 \\
    Wisconsin & - & GWO & 242 & 44 & 0.160 & 0.350 \\
     & - & PSO & 216 & 33 & 0.420 & 0.240 \\
    \bottomrule
    \end{tabular}
}
\end{table*}

\begin{table*}[t]
    \centering
    \caption{Qualitative comparison of the five different algorithms. The Accuracy, AUC, UAcc and ECE parameters are reported for different datasets}
    \label{tab: Myocardit-UA}
    \resizebox{0.97\textwidth}{!}{%
    \begin{tabular}{lllrrrr}
    \toprule
    Dataset & Feature extractor & Method & Accuracy & AUC & UAcc & ECE \\    
    \midrule
        & DenseNet121 & MCD & 97.48 & 96.76 & 95.30 & 1.22 \\
        & DenseNet121 & MCD plus PE & 97.48 & 96.76 & 95.44 & 1.16 \\
        & DenseNet121 & MCD plus GWO & 98.18 & 97.79 & 96.71 & \underline{0.67} \\
        & DenseNet121 & MCD plus BO & 98.11 & 97.59 & 96.99 & \textbf{0.58} \\
        & DenseNet121 & MCD plus PSO & 97.76 & 97.12 & 97.13 & 1.13 \\
        & ResNet50 & MCD & 97.13 & 96.74 & 94.25 & 1.97 \\
        & ResNet50 & MCD plus PE & 97.06 & 96.69 & 94.81 & 2.81 \\
        Myocardit & ResNet50 & MCD plus GWO & \textbf{98.60} & \textbf{98.30} & 98.18 & 0.80 \\
        & ResNet50 & MCD plus BO & 98.25 & 97.98 & \textbf{98.53} & 1.09 \\
        & ResNet50 & MCD plus PSO & \underline{98.39} & \underline{98.09} & \underline{98.32} & 1.16 \\
        & VGG16 & MCD & 95.87 & 95.05 & 91.59 & 1.06 \\
        & VGG16 & MCD plus PE & 96.08 & 95.45 & 93.76 & 1.15 \\
        & VGG16 & MCD plus GWO & 96.99 & 96.49 & 95.37 & 0.98 \\
        & VGG16 & MCD plus BO & 97.34 & 96.85 & 95.16 & 0.68 \\
        & VGG16 & MCD plus PSO & 97.27 & 96.75 & 95.16 & 0.94 \\
    \midrule
        & DenseNet121 & MCD & 83.62 & 83.62 & 74.50 & 3.14 \\
        & DenseNet121 & MCD plus PE & 83.88 & 83.88 & 74.38 & 1.97 \\
        & DenseNet121 & MCD plus GWO & 83.50 & 83.50 & 75.00 & 2.83 \\
        & DenseNet121 & MCD plus BO & 83.62 & 83.62 & 76.12 & 2.45 \\
        & DenseNet121 & MCD plus PSO & 83.88 & 83.88 & 75.12 & 1.89 \\
        & ResNet50 & MCD & 96.88 & 96.88 & 94.75 & 1.48 \\
        & ResNet50 & MCD plus PE & 97.00 & 97.00 & 94.75 & 1.80 \\
        Cats Vs Dogs & ResNet50 & MCD plus GWO & 97.00 & 97.00 & 96.62 & 1.43 \\
        & ResNet50 & MCD plus BO & 96.62 & 96.62 & 96.12 & 1.35 \\
        & ResNet50 & MCD plus PSO & 96.75 & 96.75 & 96.62 & 1.25 \\
        & VGG16 & MCD & 98.12 & 98.12 & 97.75 & 1.00 \\
        & VGG16 & MCD plus PE & 98.00 & 98.00 & 97.62 & \underline{0.72} \\
        & VGG16 & MCD plus GWO & \textbf{98.38} & \textbf{98.38} & \underline{97.88} & \textbf{0.68} \\
        & VGG16 & MCD plus BO & \textbf{98.38} & \textbf{98.38} & 97.62 & 0.82 \\
        & VGG16 & MCD plus PSO & 98.12 & 98.12 & \textbf{98.12} & 1.10 \\
    \midrule
        & & MCD & 98.25 & 98.61 & 99.12 & 1.95 \\
        & & MCD plus PE & 99.12 & 99.31 & 99.12 & 1.38 \\
        Wisconsin & & MCD plus GWO & 99.12 & 99.31 & \textbf{100.00} & 0.90 \\
        & & MCD plus BO & 99.12 & 99.31 & 99.12 & \underline{0.87} \\
        & & MCD plus PSO & 99.12 & 99.31 & 99.12 & \textbf{0.49} \\
    \bottomrule
    \end{tabular}
    }
\end{table*}

\begin{table*}[t]
\centering
\caption{The centers of the two distributions, $\mu_1$ and $\mu_2$, and the distance between them $Dist$ are shown for different datasets.}
\label{tab: Myocardit-Dist}
\resizebox{\textwidth}{!}{%
    \begin{tabular}{lllrrr}
    \toprule
    Dataset & Feature extractor & Method & $\mu_1$ & $\mu_2$ & Distance \\
    \midrule
        & DenseNet121 & MCD & 0.095 & 0.472 & 0.377 \\
        & DenseNet121 & MCD plus PE & 0.083 & 0.463 & 0.380 \\
        & DenseNet121 & MCD plus GWO & 0.040 & 0.380 & 0.340 \\
        & DenseNet121 & MCD plus BO & 0.036 & 0.471 & \textbf{0.435} \\
        & DenseNet121 & MCD plus PSO & 0.063 & 0.499 & \textbf{0.435} \\
        & ResNet50 & MCD & 0.139 & 0.395 & 0.256 \\
        & ResNet50 & MCD plus PE & 0.157 & 0.428 & 0.272 \\
        Myocardit & ResNet50 & MCD plus GWO & 0.020 & 0.234 & 0.214 \\
        & ResNet50 & MCD plus BO & 0.013 & 0.348 & 0.335 \\
        & ResNet50 & MCD plus PSO & 0.016 & 0.222 & 0.206 \\
        & VGG16 & MCD & 0.117 & 0.509 & 0.392 \\
        & VGG16 & MCD plus PE & 0.097 & 0.498 & 0.401 \\
        & VGG16 & MCD plus GWO & 0.047 & 0.446 & 0.399 \\
        & VGG16 & MCD plus BO & 0.067 & 0.493 & 0.426 \\
        & VGG16 & MCD plus PSO & 0.053 & 0.415 & 0.362 \\
    \midrule
        & DenseNet121 & MCD & 0.302 & 0.513 & 0.210 \\
        & DenseNet121 & MCD plus PE & 0.303 & 0.514 & 0.210 \\
        & DenseNet121 & MCD plus GWO & 0.288 & 0.508 & 0.220 \\
        & DenseNet121 & MCD plus BO & 0.293 & 0.515 & 0.222 \\
        & DenseNet121 & MCD plus PSO & 0.302 & 0.520 & 0.218 \\
        & ResNet50 & MCD & 0.106 & 0.481 & 0.375 \\
        & ResNet50 & MCD plus PE & 0.119 & 0.495 & 0.376 \\
        Cats Vs Dogs & ResNet50 & MCD plus GWO & 0.039 & 0.418 & \textbf{0.379} \\
        & ResNet50 & MCD plus BO & 0.042 & 0.420 & \underline{0.378} \\
        & ResNet50 & MCD plus PSO & 0.043 & 0.418 & 0.375 \\
        & VGG16 & MCD & 0.027 & 0.333 & 0.306 \\
        & VGG16 & MCD plus PE & 0.035 & 0.362 & 0.327 \\
        & VGG16 & MCD plus GWO & 0.023 & 0.392 & 0.369 \\
        & VGG16 & MCD plus BO & 0.023 & 0.355 & 0.331 \\
        & VGG16 & MCD plus PSO & 0.021 & 0.354 & 0.332 \\
    \midrule
        & & MCD & 0.046 & 0.604 & 0.559 \\
        & & MCD plus PE & 0.052 & 0.662 & 0.609 \\
        Wisconsin & & MCD plus GWO & 0.012 & 0.676 & \underline{0.664} \\
        & & MCD plus BO & 0.012 & 0.693 & \textbf{0.680} \\
        & & MCD plus PSO & 0.016 & 0.646 & 0.630 \\
    \bottomrule
    \end{tabular}
    }
\end{table*}

\subsection{Synthetic datasets}

In this section, we show the results achieved by the proposed framework for the synthetic dataset (Circles). The original MCD algorithm proposed by~\cite{gal2016dropout} and the framework proposed in~\cite{shamsi2021improving, shamsi2023novel} (which we will call MCD plus entropy) are used for benchmarking.\\
For consistency, we used the same architecture for all five models (MCD, MCD plus entropy, MCD plus GWO, MCD plus BO, and MCD plus PSO). Each model consists of a neural network with two hidden layers, utilizing Rectified Linear Unit (ReLU) activation functions. Notably, all models have two hidden layers containing 64 and 16 neurons, respectively.\\
Fig.~\ref{Fig:UAs} illustrates the different UAcc values obtained by varying thresholds across the algorithms: MCD, MCD plus entropy, MCD plus GWO, MCD plus BO, and MCD plus PSO. In all scenarios, proposed dual optimization method demonstrates higher UAcc values, indicating that this algorithm is more effective at gauging its confidence—specifically, it assigns higher uncertainty to incorrect predictions and lower uncertainty to correct ones.\\
Table~\ref{tab:my-table1} presents the Accuracy, AUC, UAcc, and ECE metrics for each algorithm trained on datasets with varying noise levels. The ECE metric reflects how well-calibrated the predictions are, with an ideal ECE value being zero. As shown in Table~\ref{tab:my-table1}, our dual optimization framework, incorporating GWO, BO, and PSO, consistently outperforms the MCD baseline in both traditional accuracy and UAcc, demonstrating its ability to provide more reliable uncertainty estimates. Additionally, it achieves lower ECE values across various experiments, indicating better-calibrated predictions. The integration of uncertainty into the loss function further enhances overall performance, solidifying our approach as superior to the standard MCD model across all optimization strategies.\\
Table~\ref{tab:my-table2} details the characteristics of the distributions shown in Fig.~\ref{Fig:Distributions} for each model. The variable "Dist" represents the distance between the distributions of correct and incorrect predictions. A larger Dist value indicates a better ability of the model to differentiate between correct and incorrect predictions. The values obtained for Dist suggest that our proposed frameworks achieve a larger Dist compared to other algorithms, further highlighting their superior performance.

\subsection{Real datasets} \label{section:MyocarditResults}
To validate whether our proposed algorithm can be effectively applied to real-life datasets and achieve acceptable results, we selected three different datasets: Myocardit, Cats vs Dogs, and Breast Cancer Wisconsin, which are described in more detail in Section~\ref{sec: dataset}.\\
High accuracy is a critical prerequisite for any model intended to serve as a baseline or framework for uncertainty quantification techniques, as a model with low accuracy cannot be considered reliable. Therefore, it is essential to ensure that our base models achieve acceptable accuracy on these datasets. We selected the VGG16~\cite{VGG2014}, ResNet50~\cite{he2016deep}, and DenseNet121~\cite{huang2017densely} deep neural architectures and employed a transfer learning approach. Training these models from scratch on datasets with a limited number of samples is impractical and could result in biased outcomes toward one of the classes. Additionally, training deep models from scratch is computationally inefficient, even with large datasets. To address these challenges, we utilized the pretrained weights of VGG16, ResNet50, and DenseNet121 on the ImageNet dataset to extract the most important features from the Myocardit and Cats vs Dogs datasets (the Breast Cancer Wisconsin dataset is tabular and does not require feature extraction).\\
To further optimize the data for statistical analysis, we applied Principal Component Analysis (PCA)~\cite{jolliffe2016principal} with 100 components, reducing the high dimensionality of the input data to 100 components. These features were then used as inputs to a fully connected neural network with three hidden layers, followed by a softmax classifier.\\
During the initial training of each neural network, the weights were randomly initialized. To ensure the robustness of the architecture regardless of random weight initialization, we trained the models 10 times. The distributions of Accuracy and AUC across these trials are depicted in Fig. 4. The average values for Accuracy and AUC were consistently high, confirming that the models are suitable baselines for applying uncertainty quantification techniques, such as the MCD algorithm.

In our study, the dropout hyperparameters \(P1\) and \(P2\) were selected from the interval \((0,1)\)  to determine the optimal values using the Optimizers (GWO, BO and PSO). The size of the first hidden layer, \(L1\), was chosen from the range \((64,256)\), and the size of the second hidden layer, \(L2\), was varied between \(16\) and \(64\). These ranges were chosen based on prior empirical studies indicating their effectiveness in various neural net- work architectures. The optimization process involved using three different algorithms to fine-tune these hyperparameters across multiple datasets. Table~\ref{tab:optimisedparams} presents the optimal hyperparameter configurations derived from each optimization algorithm. These configurations include the dropout rates \(P1\) and \(P2\), and the sizes of the hidden layers \(L1\) and \(L2\). To further validate the effectiveness of the estimated hyperparameters, we integrated them into our MCD model. The objective was to evaluate whether these optimized hyperparameters could enhance the MCD model’s ability to capture reliable uncertainty estimates. By doing so, we aim to ensure that the model not only achieves high predictive performance but also provides robust uncertainty quantification, which is crucial for tasks requiring high reliability. Our results indicate that the optimization algorithms significantly improved the MCD model’s performance. We observed a marked improvement in the model’s ability to quantify uncertainty, suggesting that the selected hyperparameters played a crucial role in achieving this. This improvement underscores the importance of hyperparameter optimization in developing neural networks capable of reliable uncertainty estimation.

Fig.~\ref{Fig: Myocardit} shows UAccs of five algorithms applied to 3 different feature extractions for different thresholds (It should be noted that Wisconsin is a tabular dataset and dose not require any feature extraction). The proposed algorithm significantly improves the performance of MCD compared to the others algorithms. In other words, it is better in capturing and communicating its confidence for different predictions. This characteristic is essential for all neural networks, especially when their face with rare cases or when the data is from another scanner, institution, and geographic region. Additionally, Fig.~\ref{Fig: UAThr} has been plotted to display the UAccs for a threshold of 0.5, illustrating that proposed algorithms namely GWO, Bo and PSO outperform in all cases comparing to MCD baseline utilizing different feature extractors as DenseNet121, ResNet50 and VGG16. The quantitative comparison of different algorithms are shown in Table.~\ref{tab: Myocardit-UA} for different datasets. The results of dataset show the superiority of the proposed algorithm and how it improves the current frameworks by improving the UAcc and decreasing the ECE simultaneously. In addition, the proposed algorithm improves the accuracy of the model when applying to the real datasets. This indicates that our hybrid optimization has improves both accuracy of the model and uncertainty accuracy. It should be emphasized that improving the uncertainty accuracy (UAcc) is equal to better quantifying epistemic uncertainty. The qualitative comparisons of the five algorithms are depicted in Table.~\ref{tab: Myocardit-Dist} illustrates that MCD struggles to effectively differentiate between the distributions of correctly and misclassified. However, when MCD is enhanced with dual optimization techniques namely GWO, BO, and PSO, help the two distributions being better differentiated (higher value for the distance of the centers).

Figure~\ref{Fig: com3met} presents a comparative analysis of three performance metrics (Accuracy, AUC, and UAcc) across different methods applied to the Myocardit dataset (the primary dataset in our study), with each sub-figure corresponding to a different neural network architecture (VGG16, ResNet50, and DenseNet121). The charts illustrate the performance of the five methods. The charts reveal that the dual optimization solutions generally achieve superior performance across all three metrics, underscoring the effectiveness of these enhancement techniques in improving the accuracy and robustness of neural network models.

\section{Conclusion}\label{Sec:Concl}

In this study, we introduced a novel framework that enhances MCD by integrating an uncertainty-aware loss function with advanced hyperparameter optimization technique such as GWO, Bayesian BO, and PSO. Unlike conventional MCD, which often struggles with poorly calibrated uncertainty estimates, our approach explicitly incorporates predictive entropy (PE) into the loss function. By penalizing incorrect predictions with high PE while ensuring low PE for correct predictions, our framework improves both predictive accuracy and uncertainty calibration. Extensive experiments on synthetic and real-world datasets, including Myocardit, demonstrated that our method significantly reduces the Expected Calibration Error and increases Uncertainty Accuracy. By leveraging different backbone architectures namely DenseNet121, ResNet50, and VGG16, we further validated the robustness and generalizability of our approach across diverse feature extraction settings. These improvements have significant implications for safety-critical applications, such as healthcare diagnostics, autonomous systems, and financial forecasting, where reliable uncertainty estimation is crucial for decision-making. Future work will explore additional optimization strategies and further refinements to enhance Uncertainty Accuracy, making the framework even more robust for complex, high-stakes applications.

\section{Conflict of Interest}
The authors have no conflict of interest to declare.

\section{Data availability}
The synthetic datasets (Circle) generated during the current study are from Scikit Learn library. 
The Myocardit dataset is available at Kaggle~\cite{sharifrazi_myocarditis_dataset}.
The Cats vs Dogs dataset is available at Microsoft website~\cite{microsoft_kaggle_cats_dogs}.
The Breast Cancer Wisconsin dataset can be loaded with Scikit Learn library~\cite{street1993nuclear}.

\bibliography{main}

\end{document}